\documentclass[letterpaper, 10 pt, conference]{ieeeconf}  % Comment this line out if you need a4paper

\IEEEoverridecommandlockouts                              % This command is only needed if 
\usepackage{graphicx}
\usepackage{booktabs}
\usepackage{xcolor}
\newcommand{\rev}[1]{#1}

\usepackage{booktabs}
\usepackage{array}
\usepackage{multirow}
\usepackage{amssymb}
\usepackage{amsmath}
\usepackage{pifont}

\title{\LARGE \bf
Before Humans Join the Team:\\Diagnosing Coordination Failures in Healthcare Robot Team Simulation
}

\author{Yuanchen Bai$^{1,2}$, Zijian Ding$^{3}$, Shaoyue Wen$^{4}$, Xiang Chang$^{1}$ and Angelique Taylor$^{1,2}$% <-this % stops a space
\thanks{$^{1}$Yuanchen Bai, Xiang Chang, and Angelique Taylor are with  Cornell Tech, New York, USA
        {\tt\small yb299@cornell.edu, } {\tt\small xc529@cornell.edu, }{\tt\small amt298@cornell.edu}}%
\thanks{$^{2}$Yuanchen Bai and Angelique Taylor are also with the Department of Information Science, Cornell University, Ithaca, USA}        
\thanks{$^{3}$Zijian Ding is with University of Maryland, College Park {\tt\small ding@umd.edu}}%
\thanks{$^{4}$Shaoyue Wen is with Imperial College London {\tt\small jw7525@ic.ac.uk}}%
}

\begin{document}

\maketitle
\thispagestyle{empty}
\pagestyle{empty}

%%%%%%%%%%%%%%%%%%%%%%%%%%%%%%%%%%%%%%%%%%%%%%%%%%%%%%%%%%%%%%%%%%%%%%%%%%%%%%%%
\begin{abstract}
As humans move toward collaborating with coordinated robot teams, understanding how these teams coordinate and fail is essential for building trust and ensuring safety. \rev{However, exposing human collaborators to coordination failures during early-stage development is costly and risky, particularly in high-stakes domains such as healthcare.} We adopt \rev{an agent-simulation approach in which all team roles, including the supervisory manager, are instantiated as LLM agents, allowing us to diagnose coordination failures before humans join the team.} Using a controllable healthcare scenario, we conduct two studies with different hierarchical configurations to analyze coordination behaviors and failure patterns. Our findings reveal that team structure, rather than contextual knowledge or model capability, constitutes the primary bottleneck for coordination, and expose a tension between reasoning autonomy and system stability. \rev{By surfacing these failures in simulation, we prepare the groundwork for safe human integration.} These findings inform the design of resilient robot teams with implications for process-level evaluation, transparent coordination protocols, and structured human integration. Supplementary materials, including codes, task agent setup, trace outputs, and annotated examples of coordination failures and reasoning behaviors, are available at: {https://byc-sophie.github.io/mas-to-mars/}
\end{abstract}

%%%%%%%%%%%%%%%%%%%%%%%%%%%%%%%%%%%%%%%%%%%%%%%%%%%%%%%%%%%%%%%%%%%%%%%%%%%%%%%%
\section{Introduction}

Human--robot interaction is increasingly envisioned to move beyond one-to-one interaction with individual robots toward collaboration with coordinated robot teams that assume complementary roles within shared workflows \cite{bai2026towards}. Large language model (LLM)-based multi-agent systems (MAS) provide a promising technical foundation for such robot teams, where multiple agents can support robots from capability, interaction, and implementation aspects. Specifically, MAS has emerged as an effective paradigm for solving more diverse and complex tasks (e.g., code generation \cite{qianChatDevCommunicativeAgents2024} and  debate~\cite{duImprovingFactualityReasoning2023}) compared with single-agent approaches \cite{li2024survey}.
From an interaction perspective, natural language provides an intuitive interface for coordinating robot teams, making such systems more accessible to both technical and non-technical users, including those without prior robot programming experience \cite{lu2023extracting}. For implementation, researchers have developed multi-agent frameworks, such as AutoGen \cite{wuAutoGenEnablingNextGen2023} and CrewAI \cite{crewai}, to ease the creation of MAS.

However, alongside the growing potential of robot teams, it is critical to consider the risks of system failure. Failures in robotic systems, and how they are handled and communicated to human collaborators, can significantly impact both task completion and human trust \cite{honig2018understanding,lemasurier2024templated,dossett2025trust}. 
While prior work has examined failures in human--robot interaction, it has largely focused on single-robot settings (e.g., \cite{lemasurier2024templated}). As a result, how coordination breakdowns emerge within robot teams, and what they imply for human oversight and intervention, remains underexplored.

Studying such failures in multi-agent robotic systems (MARS) introduces additional challenges. In high-stakes domains such as healthcare, physical constraints (e.g., limited robots, hardware bottlenecks, and high operational costs) make failures costly and demand efficient resource allocation \cite{taylor2020situating}. 
Coupled with strict safety and reliability requirements, these challenges call for robust coordination structures that are resilient to failure (e.g., hierarchical structure \cite{huang2024resilience}), rather than ad hoc or unstructured ones. 
This raises a critical question: whether MARS, built on MAS frameworks originally designed for virtual tasks, can meet the demands of real-world deployment.

However, existing analyses of MAS coordination patterns fall short in capturing real-world complexities. For example, \cite{cemriWhyMultiAgentLLM2025} identifies 14 failure modes across three categories, but the analysis is based on virtual task benchmarks such as math problem solving. Prior evaluations focus on task outcomes, lacking finer-grained, process-level assessments \cite{shojaee2025illusion}.
In addition, reasoning capability, an important factor shaping agent behavior, has been primarily examined at the single-agent level (e.g., \cite{liu2024mind}), with limited understanding of its impact on team-level coordination. Together, these gaps motivate our investigation into MARS coordination patterns and failure modes, toward better supporting humans who build and work alongside robot teams.

To address the absence of existing benchmarks for evaluating MARS under real-world constraints, we construct a custom, controllable scenario capable of systematically injecting critical challenges and boundary conditions—such as team-level recovery logic and hierarchical role interpretation. Among potential domains, healthcare stands out: in high-stakes tasks like emergency room onboarding, robots need to operate under resource constraints, clearly defined roles, and low fault tolerance. 
Building on this setting, we examine the performance of hierarchical MARS, built upon current state-of-the-art multi-agent frameworks, and analyze the coordination patterns that emerge across contextually grounded scenarios. 
Importantly, our goal is not to compare frameworks. Rather, we use different system configurations as controlled probes to analyze coordination behaviors and failure patterns.

\rev{Following the generative agent simulation paradigm~\cite{park2023generative}, all roles in the robot team, including the supervisory manager that would ultimately be filled by a human healthcare worker, are instantiated as LLM agents. This fully simulated approach provides a controlled testbed for characterizing coordination dynamics as a prerequisite to hybrid human-agent deployment.} Our analysis is conducted at the agent-agent level and positions humans as integral to the broader interaction loop, as system designers, collaborators, and ultimate decision-makers to whom failures may be escalated. By characterizing how failures emerge, propagate, and are resolved within robot teams \rev{before humans are introduced into the loop}, our work provides a foundation for designing more effective human--robot coordination, including how failures should be managed and communicated in collaboration with humans. \rev{This ``diagnose before deploy'' approach ensures that coordination protocols are robust and well-understood prior to human integration, reducing the risk of exposing human collaborators to preventable system failures.} In this way, we contribute toward realizing human–robot symbiosis with AI in real-world, high-stakes environments.

Our contributions include:
\begin{itemize}
    \item \textbf{Contributing Factors to Coordination Failures in Hierarchical MARS:} 
    We identify coordination failures in hierarchical MARS and investigate their dependencies on contextual knowledge, system structure, and underlying model reasoning capability. Our findings reveal that while sufficient contextual knowledge is necessary, system structure remains the bottleneck for robust coordination, and different reasoning capabilities give rise to distinct failure profiles.

    \item \textbf{Reasoning Capabilities and Coordination Trade-offs:} 
    We find that strong reasoning models exhibit more advanced planning and team orchestration in our specific scenario, but also introduce more diverse failure patterns due to their reasoning initiatives. 
    Although non-reasoning models show fewer failure patterns in our scenario, this stems not from stronger problem-solving capabilities but from a lack of deliberate reasoning that limits their autonomy and adaptability.

    \item \textbf{Design Implications for Resilient Robot Teams: } Drawing from our empirical findings, we identify three design principles for building robot teams that humans can effectively supervise and collaborate with: process-level evaluation as a foundation for surfacing otherwise invisible coordination failures, transparent coordination protocols that expose team state on demand, and structured human integration that treats human roles within the hierarchy as deliberate design choices.
\end{itemize}

\section{Related Work}

A MAS consists of multiple agents that collaborate to achieve a common goal \cite{li2024survey}. Hierarchical MAS, noted for their resilience to failure \cite{huang2024resilience}, are promising for scaling to larger and more layered agent teams for tasks of higher complexity. While some recent MAS are considered as       ``hierarchical'', they often simplify hierarchy into rigid task-handoffs, overlooking the adaptive, bidirectional structures seen in real organizations of agents. This suggests a revisit to what hierarchy entails and whether incorporating these elements could enhance coordination.

Recent work provides engineering scaffolds to support the exploration of custom MAS.
Frameworks such as Microsoft AutoGen~\cite{wuAutoGenEnablingNextGen2023}, CrewAI \cite{crewai}, and LangGraph \cite{LangGraph} represent attempts in this direction, providing modular agent construction and communication mechanisms that enable developers to flexibly configure models, tools, and interaction processes. 
These frameworks have demonstrated promising adaptability in many tasks, including crime trend analysis~\cite{fatima2025autogendrivenmultiagent}, paper reviews~\cite{li2025largelanguagemodelstrusted}, and engineering material model construction~\cite{tian2024optimizingcollaborationllmbased}. % Tools such as AutoGen Studio further simplify the threshold for developers to define agent configurations and debug collaborative workflows~\cite{dibiaAutoGenStudioNoCode2024}. 
However, once deployed in real-world physical scenarios, their originally language-driven, loosely-structured designs expose vulnerabilities due to limited physical resources and demanding delegation and report-back mechanisms.

To advance MAS applications in real environments, researchers have conducted a systematic analysis of collaborative failure modes such as agent authority overreach, role responsibility conflicts, tool invocation confusion, and feedback chain disruption~\cite{cemriWhyMultiAgentLLM2025}. 
Consequently, some research attempts to improve system robustness from the perspective of model capabilities, such as introducing reinforcement learning mechanisms~\cite{yuan2025reinforcellmreasoningmultiagent}, embedding causal or symbolic reasoning modules~\cite{wan2025remalearningmetathinkllms}, and modeling temporal dependency structures~\cite{wang2025ragenunderstandingselfevolutionllm}. 
Other work emphasizes comprehensive modeling of knowledge structures and improving external environment perception, in an attempt to improve system control over task contexts and information flow~\cite{krishnan2025advancingmultiagentsystemsmodel}.

However, even with the above enhancement mechanisms, once collaboration enters high-risk, low-tolerance real-world physical environments, the originally language-driven, loosely-structured MAS architectures still struggle to effectively address failure modes caused by their structural deficiencies~\cite{he2025comprehensivevulnerabilityanalysisnecessary, cemriWhyMultiAgentLLM2025}. 
Therefore, we focus on a high-risk, low-tolerance task—medical robotic collaboration—to systematically analyze the failure modes exposed by current mainstream multi-agent frameworks in this scenario, and propose a protocol-constrained MAS design approach to improve system task control, collaborative transparency, and structural fault tolerance.
%%%%%
% Section 3: Methodology Overview%
%%%%%
\section{Methodology}
\label{sec:method}

\begin{figure}[t]
\centering
\includegraphics[width=\columnwidth]{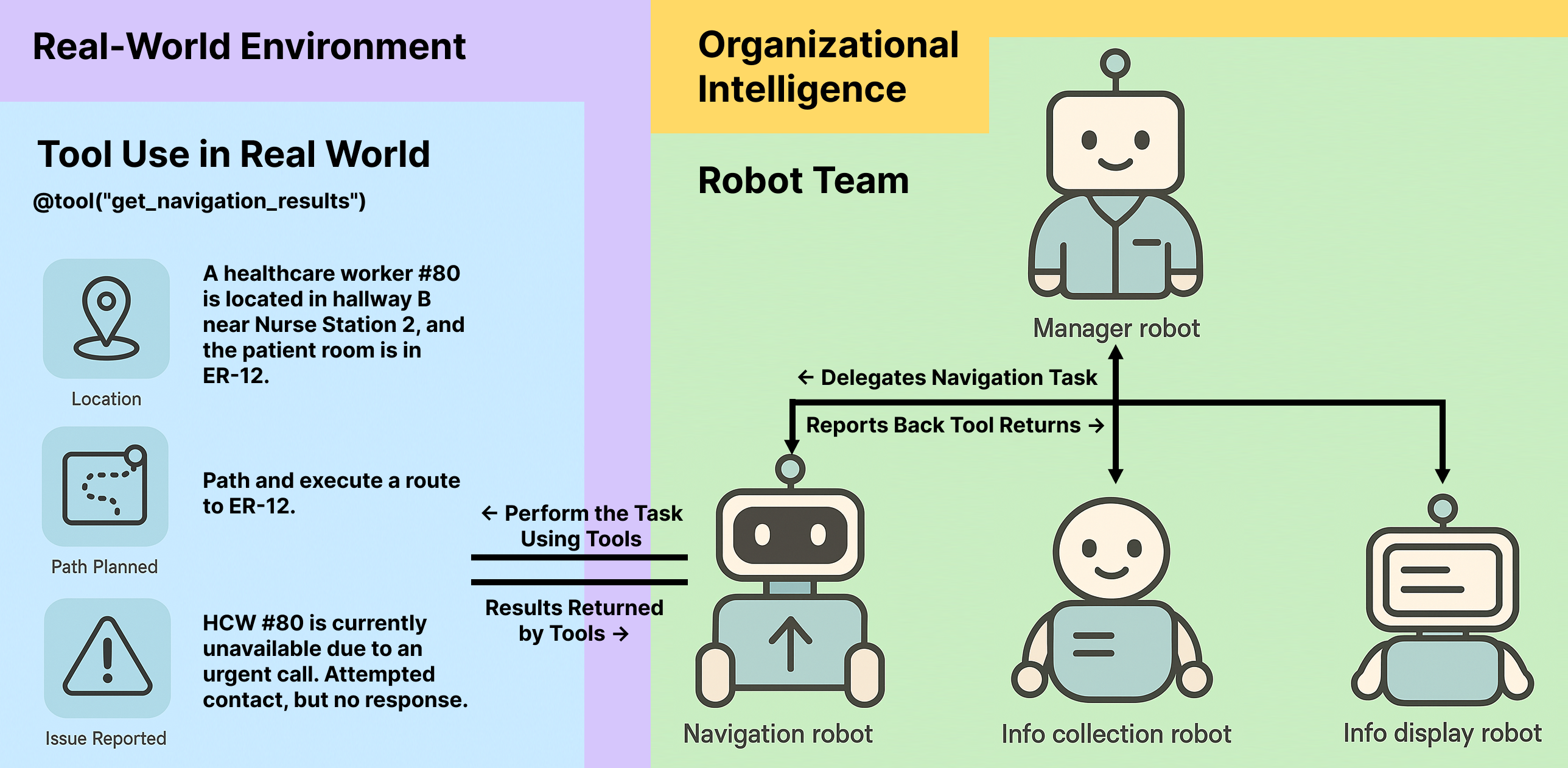}
\caption{\rev{Overview of our agent-simulated hierarchical MARS for emergency room onboarding. In this scenario, a patient arrives and a healthcare worker (HCW) must be located, credentialed, and assigned. A manager robot ($r_m$) coordinates three subordinate robots: navigation ($r_n$) locates and guides the HCW to the patient room, information collection ($r_c$) gathers HCW credentials and specialty data, and information display ($r_d$) presents team composition and generates a layout plan. Each subordinate uses a dedicated tool simulating its onboard subsystem, interprets the tool's output, and reports back to the manager. All roles are instantiated as LLM agents to diagnose coordination failures before human healthcare workers join the team.}}
\label{fig:teaser}
\end{figure}
We designed a controlled test case in a healthcare setting that simulates real-world complexity, serving as a testbed to examine how hierarchical MARS systems operate under high-stakes conditions. 
Rather than comparing frameworks, we use different system configurations as probes to examine how three factors (i.e., contextual knowledge, communication structure, and reasoning capability) shape coordination, with the goal of informing human oversight and transparency in robot team deployment.

\rev{All roles in the robot team, including the supervisory manager, are instantiated as LLM agents rather than human operators. Following the generative agent simulation paradigm~\cite{park2023generative}, this design choice is deliberate: by using agents to surface coordination failures first, we avoid exposing human healthcare workers to preventable breakdowns during early-stage system development. This allows us to systematically diagnose and address failure modes before humans join the team, establishing a foundation for subsequent hybrid human-agent integration (see Section~\ref{sec:future}).}

\begin{figure}[t]
\centering
\includegraphics[width=1.2\columnwidth]{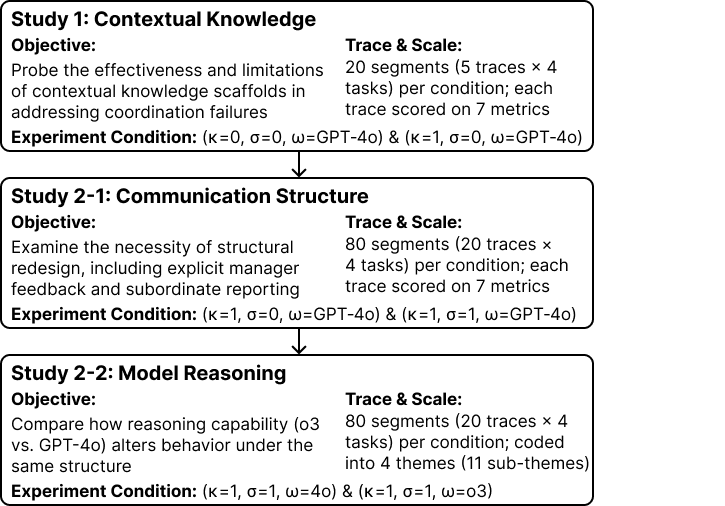} % Reduce the figure size so that it is slightly narrower than the column.
\caption{Study Overview.}
\label{fig:method_overview_summary}
\end{figure}

To illustrate our stepwise exploration of the three factors (See Figure~\ref{fig:method_overview_summary}), we define experimental settings of the studies as a configuration tuple $(\kappa, \sigma, \omega)$, where $\kappa \in \{0, 1\}$, $\sigma \in \{0, 1\}$, $\omega \in \{\text{GPT-4o},\, \text{o3}\}$.
Here, $\kappa=1$ indicates the inclusion of contextual and procedural knowledge, while $\kappa=0$ corresponds to its absence. 
$\sigma=1$ denotes an enhanced communication structure, and $\sigma=0$ reflects its absence. 
$\omega$ specifies the underlying model, either GPT-4o-2024-08-06 \cite{4o} or o3-2025-04-16 \cite{o3}, respectively representing non-reasoning and reasoning models.

\paragraph{MARS Framework, Robots, Tools and Tasks Setup.}

\begin{table}[t]
  \centering
  \caption{MARS roles and corresponding tools.}
  \label{tab:robot_tools}
  
  \begin{tabular}{>{\raggedright\arraybackslash}p{0.3\columnwidth}>{\raggedright\arraybackslash}p{0.6\columnwidth}}
    \toprule
    \textbf{Robot Role} & \textbf{Robot Tool (Simulated Subsystem)} \\
    \midrule
    $r_n$: Locates HCWs and guides them to assigned rooms.  
      & $u_n$: Simulates internal navigation systems including location tracking, path planning, and staff communication. \\
    \midrule
    $r_c$: Gathers HCW credentials and specialty data.  
      & $u_c$: Simulates the onboard interface to collect structured identity and specialty information. \\
    \midrule
    $r_d$: Presents data and generates layout plan.  
      & $u_d$: Simulates internal systems to query the institutional database to retrieve and display team roles and composition details. \\
    \midrule
    $r_m$: Orchestrates the team.
      & No assigned tools beyond built-in coordination functions (e.g., delegation). \\
    \bottomrule
    \end{tabular}
\end{table}

MARS comprises three core components—\textbf{robots role} ($R$), \textbf{tasks} ($T$), and \textbf{tools or utilities} ($U$).  While we retain the term ``tool'' for consistency with LLM-agent frameworks terminology, these tools simulate the subsystems of robots in MARS.
Let $R = \{r_m, r_n, r_c, r_d\}$ denote the set of four robots: a manager ($r_m$) and three subordinates—navigation ($r_n$), information collection ($r_c$), and information display ($r_d$) robots. Each subordinate robot is uniquely equipped with a corresponding tool from the tool set $U = \{u_n, u_c, u_d\}$. We define a one-to-one mapping $\psi: \{r_n, r_c, r_d\} \rightarrow U$ such that: $ \psi(r_n) = u_n, \psi(r_c) = u_c, \psi(r_d) = u_d.$
The manager robot $r_m$ is not assigned any task-execution tools. Instead, it relies solely on built-in coordination functions provided by the framework (e.g., delegation). This reflects its intended role as a high-level planner and leader, rather than an executor of low-level tasks handled by subordinate robots.

The overall task set is denoted as $T = \{ \tau_n, \tau_c, \tau_d, \tau_{\text{ref}} \}$. $T$ reflects a minimal but representative coordination workflow adapted from acute-care onboarding. It comprises three execution tasks—navigation ($\tau_n$), information collection ($\tau_c$), and information display ($\tau_d$)—which are expected to be completed by their corresponding subordinate robots, and a reflection task ($\tau_{\text{ref}}$) (i.e. to reflect on the overall process and summarize outcomes and lessons learned), which is expected to be handled solely by the manager. We impose a strict precedence relation $\tau_n \prec \tau_c \prec \tau_d \prec \tau_{\text{ref}}$, meaning that each downstream task may start only after its immediate predecessor has been marked \textit{Success} or its failure has been resolved by the manager. This structure reflects the high-stakes nature of real-world settings such as healthcare, where downstream actions (e.g., data interpretation or decision-making) must wait until upstream requirements—such as staff arrival or identity confirmation—are satisfied. Each robot uses and can only use its own tool to complete its assigned task, and then reports the result to the manager. The manager is responsible for validating the task completion outcomes and determining follow-up actions, such as retry or escalation.
Table \ref{tab:robot_tools} shows the description of robots' roles and tools.

\paragraph{Test Case Design.}
\label{sec:testcase}

We design a test case \(\Phi = (O,\, \mathcal{P},\, \{\delta_j\})\), comprising a predefined observation \(O\), prompt pack \(\mathcal{P}\), and scripted tool returns \(\delta_j\) (See Table \ref{tab:test_case_summary}). \(O\) encodes situational context across all tasks, including observable cues and ambient constraints that agents may use to infer what needs to be done. In practice, \(O\) simulates the environment as perceived by robots (e.g., whether upstream failure is resolved) which serve to trigger appropriate tasks and ground their execution. Thus, \(O\) plays a dual role: it provides contextual justification for why a task becomes relevant and offers cues needed to reason about how to execute it. \(\mathcal{P}\) includes task descriptions, robot roles and any other contextual knowledge. \(\delta_j\) is a structured, tool-specific output (e.g., a planned avigation path) returned by a robot subsystem. These outputs require \textit{further interpretation} by the robot to assess whether the task has succeeded or failed. When a failure is inferred, the robot must \textit{escalate} the issue by reporting it to the manager for high-level coordination or recovery. This setup offers a controlled yet realistic environment for eliciting agent behaviors in the face of both expected and unexpected outcomes.

\begin{table}[t]
\centering
\caption{Observations, corresponding tasks, and expected outcomes for our MARS studies.}
\label{tab:test_case_summary}
\begin{tabular}{p{0.4\columnwidth} p{0.5\columnwidth}}
\toprule
\multicolumn{2}{l}{\textbf{Observation $o_n$: Patient Arrival but HCW Unavailable}} \\
\midrule
\textbf{Corresponding Task} & \textbf{Expected Outcome} \\
\cmidrule(lr){1-2}
$\tau_n$: Navigate the healthcare worker to the assigned patient room  
& Detect and escalate task failure due to HCW unavailability; perform failure handling \\
\midrule

\multicolumn{2}{l}{\textbf{Observation $o_c$: HCW Reassigned \& Collection Begins}} \\
\cmidrule(lr){1-2}
$\tau_c$: Collect identity and specialty data from the newly assigned HCW \#90  
& Recognize that upstream issue has been resolved; successfully retrieve HCW information without issue \\
\midrule

\multicolumn{2}{l}{\textbf{Observation $o_d$: Team Info Collected \& Layout Updated}} \\
\cmidrule(lr){1-2}
$\tau_d$: Get updated data and generate a visual layout plan  
& Display correct team information as prompted and produce a layout plan reflecting current assignments \\
\midrule

\multicolumn{2}{l}{\textbf{Observation $o_\text{ref}$: Post-Task Reflection Report}} \\
\cmidrule(lr){1-2}
$\tau_{\text{ref}}$: Generate a post-hoc summary of team performance  
& Produce an accurate report summarizing outcomes, reasoning, and lessons learned across prior tasks \\
\bottomrule
\end{tabular}
\end{table}

To characterize the complexity of MARS coordination, we present a structured decision loop to illustrate how robot behavior at each step depends on multiple factors: 
At each time step \(t\), a robot \(r\) selects an action \(a_t\) using a policy \(\pi_\omega\) instantiated by the underlying model \(\omega\), conditioned on the current observation by the robot (e.g., via sensors) \(O_t\), cumulative interaction history \(H_t\), task assigned \(\tau\), and prompt pack \(\mathcal{P}\):
$a_t = \pi_\omega(r, \tau, O_t, H_t, \mathcal{P}).$
The action \(a_t\), once executed, updates the environment and leads to new states for future decision steps: $(O_t, H_t) \xrightarrow{a_t} (O_{t+1}, H_{t+1})$.

% We expect our test case to reveal diverse coordination patterns and serve as a rich testbed for analyzing MARS.

\paragraph{Challenges in High-Stakes Real-World Tasks with Robot Team.}
As no established metrics exist for hierarchical MARS, we begin by identifying seven dimensions where real-world constraints introduce distinct operational considerations in MARS, forming the basis for our evaluation criteria and test case design. These criteria reflect coordination challenges that—while some may potentially present in other MAS contexts—have not been systematically examined. In our high-stakes healthcare setting, such challenges become especially pronounced: \textit{1) Agent Characteristics} shift from ephemeral modules to persistent, embodied team members requiring stable identity and accountability. \textit{2) Agent Configuration} becomes role-based, with agents managing a coherent cluster of functions over time. \textit{3) Role Boundaries \& Constraints} are tightly scoped due to hardware and authority control. \textit{4) Traceability \& Accountability} is heightened, as failures need to be linked to specific agents and actions. \textit{5) Consequence of Upstream Failures} increases, since downstream tasks frequently depend on upstream success without fallback mechanisms. \textit{6) Tool Access \& Modularity} is shaped by how tools are embedded within individual robot systems, making them less interchangeable and more tightly bound to agent roles compared to abstract, API-like tools. \textit{7) Definition of Success} becomes holistic—dependent not only on task completion but also on correct sequencing, reporting, and compliance with role expectations.

\section{Study 1: Evaluation of Hierarchical MARS Coordination}

Study 1 serves as an initial diagnostic to screen for typical coordination failures in hierarchical MARS. By providing rich contextual and procedural knowledge ($\kappa=0\rightarrow1, \sigma=0, \omega=\text{GPT-4o}$), we examine which failures can be addressed through knowledge alone and which persist—thereby revealing potential structural bottlenecks for further investigation (see Figure \ref{fig:method_overview_summary}).
For evaluation, we ran 5 traces (a trace refers to a full run of the four tasks) per condition.

\subsection{Experiment Setup}

\subsubsection {Hierarchical Structure Setup.}
We built on the CrewAI \cite{crewai} framework using its hierarchical mode--described in the documentation as a structure that ``simulates traditional organizational hierarchies for efficient task delegation and execution''. CrewAI also features a Knowledge Base (KB) that provides agents with ``a reference library they can consult while working.''
CrewAI is well-suited for our research because these features make it a natural starting point to probe context-grounded behaviors and to identify which coordination challenges can, or cannot, be resolved through contextual knowledge alone.

\subsubsection{Knowledge Base as Contextual Intervention.}
\label{sec:KB}

We developed a KB with contextual or procedural knowledge as a shared resource analogous to organizational documentation that ground MARS team behavior and decision-making.
The KB defines five knowledge key points, including 1) \textit{tool access rules} through a tool-robot mapping and real-world functions to prevent tool misuse, 2) \textit{role-specific responsibilities} to ensure that robots adhere to defined scope of responsibilities, 3) \textit{task success and failure criteria} to help MARS determine whether to proceed, retry, or escalate, reducing false completions, 4) \textit{environmental cue grounding} to enable MARS to learn how to interpret real-world triggers (e.g., ID scans) for initiating appropriate actions, and 5) \textit{task execution and recovery workflow} for clear procedural steps, including escalation paths, enabling MARS to adapt effectively to failure.

\paragraph{Metrics.} Our evaluation goes beyond task outcomes to capture process-level dynamics, particularly within the hierarchical structure. We assess the performance of MARS at both the manager and subordinate levels. All metrics are evaluated using a rubric-based scheme with three levels: 0 (criterion not met), 0.5 (partially met), and 1 (fully met). Full task-specific rubrics are provided in the Supplementary Materials.

At the \textbf{manager level}, we consider four metrics:
\begin{enumerate}
  \item Delegation Accuracy ($M_1$): Whether $r_m$ delegates tasks to the correct robots, based on their role and tool access.
  \item Task-Completion Judgment ($M_2$): Whether the manager correctly assesses the success or failure of each task.
  \item Issue Handling ($M_3$): Whether $r_m$ detects and responds to reported issues in a timely manner.
  \item Reflection Quality ($M_4$): Whether $r_m$ reflects on the task outcomes and the relevant lessons learned captured.
\end{enumerate}

At the \textbf{subordinate robot level}, we track three metrics:
\begin{enumerate}
    \item Tool Usage ($M_{5}$): whether the agent uses the correct and accessible tool for the assigned task.
    \item Local Reasoning ($M_{6}$): whether the agent correctly executes its assigned responsibility and properly interprets the tool’s output based on the prompt.
    \item Report Compliance ($M_{7}$): whether the agent correctly reports the task execution result back to $r_m$.
\end{enumerate}

\paragraph{Scoring.} Let $\Lambda_\tau$ denote the metric set applicable for a task $\tau$ and $\Delta_\text{trace}$ denote the set of task–metric pairs present in a trace,
where $\tau$ indexes a task and $k$ indexes an evaluation metric:
$
  \Delta_\text{trace} = \bigl\{\,(\tau,k)\mid \tau\in \text{trace},\; M_k\in\Lambda_\tau \bigr\}.
$
Each pair $(\tau,k)\in\Delta_\text{trace}$ is independently scored
$s_{\text{trace},(\tau,k)}=\rho\!\bigl(\text{output}_{\text{trace},\tau},\,M_k\bigr)$,
where $\rho(\cdot,M_k)$ maps the model output to $\{0,0.5,1\}$ under metric $M_k$ via the rubric discussed previously.
The normalized success rate for a trace is the arithmetic mean of these scores:
$
\text{SR}_\text{trace} \;=\; \operatorname{mean}_{(\tau,k)\in\Delta_\text{trace}} s_{\text{trace},(\tau,k)}.
$

\subsection{Results}

Table~\ref{tab:metric_comparison} shows the performance of MARS with
and without a KB. A detailed KB leads to an increase in overall success rate—from an average of 45.29\% to 72.94\%.
Overall, KB helps increase performance across most of the seven metrics. Five metrics showed apparent improvement in mean score: delegation accuracy and reflection quality at the manager level, and tool usage, local reasoning, and report compliance at the subordinate level. 
In contrast, the KB has a minor impact on task completion and issue handling measures. 
Task completion judgment, achieved near-perfect accuracy with or without the KB, indicating this ability was already strong. 
Issue handling indicates even with detailed failure handling instructions and manager role-based emphasis (e.g., reinforcing that the manager should respond to reported issues), proactive failure handling remains lacking.

\begin{table}[h!]
  \centering
  \caption{Comparison of average performance metrics with and without a Knowledge Base (KB).}
  \label{tab:metric_comparison}
  \begin{tabular}{
    >{\raggedright\arraybackslash}p{3.2cm}
    >{\centering\arraybackslash}p{1.5cm}
    >{\centering\arraybackslash}p{1.5cm}
  }
    \toprule
    \textbf{Metric} & \textbf{$\kappa{=}0$} & \textbf{$\kappa{=}1$} \\
    \midrule
    Avg Success Rate (\%) & 45.29 & \textbf{72.94} \\
    \midrule
    Delegation Accuracy & 0.33 & \textbf{0.73} \\
    Task Completion & 0.93 & \textbf{0.97} \\
    Issue Handling & 0.00 & 0.00 \\
    Reflection Quality & 0.30 & \textbf{0.80} \\
    \midrule
    Tool Usage & 0.33 & \textbf{0.67} \\
    Local Reasoning & 0.40 & \textbf{0.73} \\
    Report Compliance & 0.47 & \textbf{0.77} \\
    \bottomrule
\end{tabular}
\end{table}

Our failure mode analysis reveals that five critical failure modes persist even with a detailed KB. Table~\ref{tab:failure_modes} summarizes these failure modes and reports their frequency across the 10 traces (n/10).
Across the 10 traces, eight traces exhibit \textit{hierarchical role misalignment}, where the manager completes tasks intended for its subordinates while delegating tasks that fall under its own leadership responsibilities; all 10 traces contain at least one \textit{tool access violation}, where a tool is used by a robot without the designated permission; and all traces show \textit{lack of in-time handling of failure reports}, in which the manager fails to provide timely alternative solutions or escalate reported problems. Four traces display \textit{noncompliance with prescribed workflows}, where agents either interpret instructions differently or ignore the given prompts; and two traces reveal \textit{bypassing or false reporting of task completion}, where the manager claims a task is complete without actually performing the required actions.
Each of these failure types corresponds to at least one KB section that has provided detailed guidelines.
Notably, even when an extensive failure recovery protocol is provided, the team consistently fail to detect, escalate, or recover from critical errors. This suggests that the bottleneck does not lie in information availability, but rather in structural limitations that inhibit timely communication and intervention.

\section{Study 2: Structural Redesign \& Model Comparison}

Study 2 extends the contextual knowledge focus of Study 1 by investigating two additional factors: communication structure and model reasoning. In Study 2-1, we address failure-handling gaps by adding explicit bidirectional communication between manager and subordinates. In Study 2-2, we examine how reasoning capacity influences coordination under the improved structure.
CrewAI's hierarchical mode helps surface coordination challenges but lacks transparency and control, prompting our transition to AutoGen, which offers a more customizable design space.

\begin{table}[t]
  \centering
  \caption{Study 1 summary of failure modes and examples.}
  \label{tab:failure_modes}
  \begin{tabular}{
    >{\raggedright\arraybackslash}p{0.34\columnwidth}
    >{\raggedright\arraybackslash}p{0.52\columnwidth}
  }
    \toprule
    \textbf{Failure Mode} & \textbf{Observed Example} \\
    \midrule
    Hierarchical role misalignment (8/10) & 
    $r_m$ completes the $\tau_c$ without delegating to the $r_c$ \\
    \midrule
    Tool access violations (10/10) & 
    $r_m$ uses the $u_c$, which should only be accessible to the $r_c$ \\
    \midrule
    Lack of in-time handling of failure reports (10/10) & 
    When the $r_n$ reports an issue (e.g., HCW unavailable), $r_m$ fails to offer alternative solutions or escalate the issue \\
    \midrule
    Noncompliance with prescribed workflows (4/10) & 
    $r_m$ pre-fetches display information and gives it as context to $r_d$, which then redundantly uses the tool to retrieve the same data again \\
    \midrule
    Bypassing or false reporting of task completion (2/10) & 
    $r_m$ claims the $\tau_m$ is complete without actually generating a report (e.g., ``Action: None (compiling the final report)'') \\
    \bottomrule
\end{tabular}
  
\end{table}

\subsection{Experiment Setup} 

\subsubsection{2-1: Hierarchical Structure Redesign.}
We implement a hierarchical structure using AutoGen's \cite{wuAutoGenEnablingNextGen2023} ``SelectorGroupChat'', where the selector decides which agent is assigned to a specific task. 
We implement two improvements to the communication structure: 
(1) \textit{Enabling proactive manager feedback:} To ensure that $r_m$ actively monitors progress, we force $r_m$ to provide timely feedback after each task execution, via a selector function.
(2) \textit{Enabling subordinate-level interpretation and report-back:} To allow subordinate agents to reflect on the outcomes of their tool usage (i.e., outputs returned from robot subsystems), we activate the `reflect\_on\_tool\_use' setting. This enables subordinates to further interpret whether the tool return indicates task success or failure, and to compose a report-back message to $r_m$.
For evaluation, we ran 80 segments (20 traces, 4 tasks per trace) using GPT-4o and re-applied the seven evaluation metrics from Study~1.

\subsubsection{2-2: Model Reasoning Comparison.}

To analyze how model reasoning influences coordination, we conducted an additional 80 segments (20 traces, 4 tasks per trace) using a strong-reasoning model (o3), compared to GPT-4o. Both of the models are released by OpenAI. Combined with the expanded communication structure introduced in Study 2-1, this setup led to more diverse and complex coordination behaviors.
To capture these nuanced patterns, we moved beyond the discrete 0/0.5/1 scoring rubric and adopted the Grounded Theory approach \cite{glaser2017discovery}, a qualitative method that facilitates theory development from empirical data. This methodology has also been used in recent MAS studies that analyze model traces (e.g. \cite{cemriWhyMultiAgentLLM2025}).
The first author developed the initial set of codes. To ensure reliability, the codes were collaboratively reviewed by four authors—all with experience in MAS-who iteratively resolved inconsistencies and refined the coding scheme until consensus was reached.

\subsection{Results}
\subsubsection{Structural Redesign Effectiveness for Failure Handling.}

The average success rate is 88.97\%. We observe strong performance across all seven metrics: delegation accuracy 88\%, task completion 88\%, issue handling 90\%, reflection quality 95\%, tool usage 90\%, local reasoning 86\%, and report compliance 90\%. 18/20 GPT-4o traces achieve consistent scores of 1 or 0.5 across all dimensions while in 2/20, the manager fails to delegate any task and hallucinates the entire workflow. Notably, issue handling, which was completely absent in Study 1, now shows marked improvement, with $r_m$ proactively generating alternative plans or escalating unresolved issues to human supervisors. In our setup, human escalation refers to invoking a higher-level external entity (beyond the robot team) when internal resolution proves insufficient.
While we acknowledge that some behavioral differences may stem from the underlying framework, the marked improvement in failure handling can be largely attributed to our structural intervention. The observed behavioral improvements, i.e., managers providing real-time feedback and subordinates proactively reassessing outcomes and reporting anomalies, closely align with the bidirectional communication mechanisms introduced in our design.
This suggests that addressing structural bottlenecks, such as communication flow, is essential for resolving persistent coordination failures like failure handling.

\subsubsection{Reasoning Trade-offs.}
We identify four major themes in MARS coordination patterns, each comprising several sub-themes (Table~\ref{tab:integrated_themes}). To contextualize these sub-themes, we annotate each with `\ding{51}' or `\ding{55}' to indicate whether its implications are positive or negative within our test scenario. We also report the frequency of each sub-theme across 20 traces for both GPT-4o and o3. We find distinct behavioral profiles which underscore trade-offs between reasoning and non-reasoning models:

\emph{1) Planning Granularity \& Execution Alignment: } 
o3 demonstrates fine-grained, step-by-step planning (1.1) in all 20 traces, often incorporating time thresholds and conditional logic (e.g., if-else), whereas GPT-4o generates only high-level, general plans, highlighting o3's greater initiative in decomposing tasks.
o3 also proactively anticipates downstream actions (1.2) in 6 traces, compared to only 1 for GPT-4o, further underscoring its planning ability.
However, this strength comes at a cost: o3 deviates from prompt instructions (1.3) in 14 traces, significantly more than GPT-4o (4 traces), reflecting o3's greater tendency to override expected procedures with its own internal logic.

\emph{2) Task \& Organizational Role Interpretation: }
o3 exhibits stronger awareness of team roles, initiating cross-role coordination (2.1) in 10 traces versus 1 for GPT-4o, indicating o3's ability to coordinate the robot team to improve task execution.
Both models reject tasks outside their scope of responsibility (2.2), but o3 does so slightly more often (5 traces) than GPT-4o (4 traces).
o3 triggers human intervention (2.3) in 13 traces, compared to 5 for GPT-4o. When neither model escalates, o3 tends to attempt new solutions, whereas GPT-4o often stalls in unproductive self-reflection.

\emph{3) Communication Robustness \& Format Compliance: }
In 9 traces, o3 ignores explicit instructions from the manager and refuses to adjust its output accordingly (3.1). In 11 traces, o3 produces outputs that deviate entirely from the expected schema (3.2). While GPT-4o occasionally produces minor formatting mismatches, it does not exhibit such schema-breaking behavior.
However, o3 demonstrates stronger auditing behavior (3.3): in the 11 traces where formatting errors occur, the manager actively verifies output completeness and explicitly flags missing or malformed fields-a level of diligence absent in GPT-4o.
This shows that while reasoning enables more rigorous self-auditing, its occasional refusal to comply with feedback limits the system's ability to recover from detectable failures.

\emph{4) Task Termination \& Verification: } 
In 20 traces, o3 repeatedly re-executes tasks without providing justification (4.1)—such as generating multiple reflection reports even after a complete one has already been produced—compared to just 1 such instance in GPT-4o. In 11 traces, o3 engages in elaborate domain reasoning to resolve issues; however, this reasoning is often unverified by actual tool returns, leading to unverifiable or factually inaccurate assertions of success (4.2). In contrast, such behavior occurs in 2 traces for GPT-4o, indicating that while elaborate reasoning enables stronger problem-solving ability, it can also increase the risk of ungrounded execution.

% These observations highlight the need for more adaptive and scenario-grounded evaluation paradigms, ones that stress-test systems under boundary conditions and varying real-world expectations. As recent findings show that reasoning can, in some cases, degrade performance on specific tasks \cite{liu2024mind}, further supporting our call for task-dependent evaluation paradigms capable of capturing the complexity and coordination demands of real world.

\section{Discussion \& Future Work}

\subsection{Technical Evaluation of Hierarchical MARS}
Through two studies using a custom test case in a healthcare MARS scenario, we investigated coordination failures and behavioral trade-offs across non-reasoning and reasoning models—highlighting a deeper tension between autonomy and stability in deploying MARS in real-world settings.
\textbf{Coordination Failures:} Though contextual knowledge is necessary to improve procedural execution, structure is the bottleneck for performance. It is essential to ensure that agents are not only given clear guidance on what tasks to perform and how, but also structurally enabled to carry out intended behaviors.
\textbf{Reasoning Trade-off:} Strong reasoning does not guarantee stability of coordination behaviors. o3 demonstrates strong problem-solving capability in orchestrating the team and generating detailed plans, but can become trapped by its own reasoning logic—such as repeatedly requesting information explicitly marked as unnecessary in the prompts, or generating redundant reflection reports after a sufficient one has already been generated. 
These overthinking behaviors align with prior observations \cite{shojaee2025illusion} that reasoning models can continue exploring alternatives even after reaching correct solutions.
In contrast, GPT-4o, exhibits shallower reasoning than o3, but can still behave unexpectedly. 
For instance, it asks the information collection robot to address a navigation failure, resembling a desperate attempt to address a perceived impasse without a grounded strategy.
Both models exhibit instability—o3 due to overthinking, and GPT-4o due to lack of deliberative reasoning. 
This suggests that instability does not stem from how much reasoning occurs, but from whether the reasoning style can be properly understood, constrained, aligned, and grounded. 
Deploying models—regardless of their level of reasoning—to operate with a degree of autonomy requires systematically understanding and managing the challenges of stability.

\subsection{Build and Collaborate with Resilient Robot Team}
Our findings point toward a broader design challenge: robot teams in high-stakes environments must be legible not only to each other, but to the humans who supervise and collaborate with them. To build and work with a resilient robot team requires deliberate design across three aspects. 
\textbf{1) Process-Level Evaluation.} Outcome-level metrics are insufficient for understanding robot team behavior-failures such as silent escalation breakdowns and unverified task completion are only detectable through fine-grained, trace-level inspection. Building human-compatible robot teams requires strict process-level evaluation before deployment.
\textbf{2) Transparent Coordination Protocols.} Beyond evaluation, deployed robot teams must be designed to expose their internal coordination state on demand, for example, structured failure report templates that encode what failed, which agent, and what recovery was attempted, and process-level update mechanisms that allow external observers to query team state without interrupting execution.
\textbf{3) Structured Human Integration.} Human roles within robot teams must be specified with the same rigor as robot roles: at which hierarchical layer does a human operator sit, under what conditions should the team surface a decision to a human, and what information does that human need to intervene effectively? These are coordination design choices that determine whether human oversight is genuinely integrated or merely reactive.

\rev{\subsection{From Agent Simulation to Human-Agent Teams}
\label{sec:future}
All team roles in our studies, including the supervisory manager, are played by LLM agents~\cite{park2023generative}. This fully simulated setup enables controlled analysis of coordination patterns as a necessary first step before deploying robot teams alongside human collaborators. The failures we identify (hierarchical role misalignment, escalation breakdowns, reasoning-compliance tension) are challenges that human supervisors will inevitably encounter.

Our findings inform a hybrid architecture in which human and agent roles are structurally interchangeable. For instance, the manager could be replaced by a human charge nurse who receives failure reports and makes escalation decisions, while subordinate robots handle execution. Future work will investigate hybrid human-robot teams where human healthcare workers selectively assume roles within the hierarchy, enabling us to study how coordination dynamics shift when humans and agents collaborate as interchangeable teammates.}

% Force all pending floats out before references
\clearpage

\begin{table}[t]
  \caption{Study 2 theme-level analysis with trace counts for GPT-4o and o3. Bold = more desirable.}
  \centering
  \scriptsize
  \begin{tabular}{
    >{\raggedright\arraybackslash}p{1.6cm}
    >{\raggedright\arraybackslash}p{3.8cm}
    >{\centering\arraybackslash}p{0.4cm}
    >{\centering\arraybackslash}p{0.4cm}
  }
    \toprule
    \textbf{Sub-theme} & \textbf{Description} & \textbf{4o} & \textbf{o3} \\
    \midrule
    \multicolumn{4}{l}{\textbf{Theme 1: Planning \& Execution Alignment}} \\
    \cmidrule(lr){1-4}
    1.1\ Step-by-Step Planning \ding{51} & Multi-layered conditional guidance with time thresholds and if-else logic & 0 & \textbf{20}\\
    1.2\ Downstream Anticipation \ding{51} & Anticipate and initiate downstream tasks without prompting & 1 & \textbf{6}\\
    1.3\ Prompt Deviation \ding{55} & Ignore explicit instructions in favor of internal logic & \textbf{4} & 14 \\
    \midrule
    \multicolumn{4}{l}{\textbf{Theme 2: Role Interpretation}} \\
    \cmidrule(lr){1-4}
    2.1\ Cross-role Collab. \ding{51} & Invoke cross-role resources to orchestrate collaboration & 1 & \textbf{10}\\
    2.2\ Task Rejection \ding{51} & Clarify role scope and refuse out-of-scope tasks & 4 & \textbf{5}\\
    2.3\ Human Escalation \ding{51} & Escalate to human when internal resolution fails & 5 & \textbf{13}\\
    \midrule
    \multicolumn{4}{l}{\textbf{Theme 3: Communication \& Compliance}} \\
    \cmidrule(lr){1-4}
    3.1\ Refusal to Coord. \ding{55} & Refuse manager feedback, turning recoverable errors fatal & \textbf{0} & 9 \\
    3.2\ Missing Output \ding{55} & Correct tool use but fail to produce structured output & \textbf{0} & 11\\
    3.3\ Manager Auditing \ding{51} & Detect missing fields and flag output errors & \textbf{0} & 11\\
    \midrule
    \multicolumn{4}{l}{\textbf{Theme 4: Termination \& Verification}} \\
    \cmidrule(lr){1-4}
    4.1\ Unjustified Repeat \ding{55} & Re-execute tasks without justification & \textbf{1} & 20\\
    4.2\ Unverified Inference \ding{55} & Elaborate reasoning unverified by tool returns & \textbf{2} & 11\\
    \bottomrule
\end{tabular}
  \label{tab:integrated_themes}
\end{table}

\bibliographystyle{IEEEtran}
%\bibliography{references_zotero}
\bibliography{references.bib}
\end{document}